\documentclass[fleqn,10pt]{wlscirep}
\usepackage[utf8]{inputenc}
\usepackage[T1]{fontenc}
\usepackage{tabularx}
\usepackage{longtable}
\usepackage{float}

\title{Project Rosetta: A Childhood Social, Emotional, and Behavioral Developmental Ontology}

\author[1,*]{Alyson Maslowski}
\author[1]{Halim Abbas}
\author[1]{Kelley Abrams}
\author[1]{Sharief Taraman}
\author[1]{Ford Garberson}
\author[1]{Susan Segar}
\affil[]{Cognoa, Inc., Palo Alto, 94306, USA}

\affil[*]{alyson@cognoa.com}

\begin{document}
\begin{abstract}
There is a wide array of existing instruments used to assess childhood behavior and development for the evaluation of social, emotional and behavioral disorders. Many of these instruments either focus on one diagnostic category or encompass a broad set of childhood behaviors. We built an extensive ontology of the questions associated with key features that have diagnostic relevance for child behavioral conditions, such as Autism Spectrum Disorder (ASD), attention-deficit/hyperactivity disorder (ADHD), and anxiety, by incorporating a subset of existing child behavioral instruments and categorizing each question into clinical domains. Each existing question and set of question responses were then mapped to a new unique Rosetta question and set of answer codes encompassing the semantic meaning and identified concept(s) of as many existing questions as possible. This resulted in 1274 existing instrument questions mapping to 209 Rosetta questions creating a minimal set of questions that are comprehensive of each topic and subtopic. This resulting ontology can be used to create more concise instruments across various ages and conditions, as well as create more robust overlapping datasets for both clinical and research use.


\end{abstract}

\flushbottom
\maketitle
%
%
\thispagestyle{empty}

\newpage
\section*{Introduction}

According to the 2011-12 National Survey of Children's Health, researchers found that approximately 1 out of 7 U.S. children aged 2 to 8 years were reported to have a diagnosed mental, behavioral, or developmental disorder \cite{CDC:2016}.  Identifying and addressing these concerns is of great importance so that interventions can start as early as possible when they have the greatest potential for improved lifelong outcomes. There are many instruments available to clinicians for the early identification of mental, behavioral, or developmental disorders. However, many of these screening tools are widely used for predicting caseness, i.e. to identify individuals who are at high risk of having at least one psychiatric disorder, while others are primarily used for identifying a specific disorder. This presents many challenges for clinicians regarding which screening tools to use for making diagnostic decisions and detecting significant effects for appropriate therapeutic interventions \cite{Screening:2017}.

The undertaking of project Rosetta was to address these challenges by creating a comprehensive ontology of concepts that have diagnostic relevance for child behavioral conditions, which can be used as a resource for child mental, behavioral, and developmental health diagnosis and treatment. This ontology can be used to create more concise instruments allowing for identification of maximally predictive minimal subsets of features associated with each behavior. Rosetta will also be able to be used to create a virtual diagnostic instrument that covers more patients in a uniform way by having overlap between existing instruments and corresponding mappings that was not present before. This will make it possible to build improved machine learning algorithms that require dense datasets for creating innovative diagnostic tools.

The first generation of project Rosetta utilized eight instruments as inputs, as shown in Table \ref{tab:Instruments}. These were chosen because they cover a wide range of child behaviors and diagnoses and were identified by subject matter experts to create a robust ontology. Some of these instruments are relatively time-consuming assessments that have to be completed by trained professionals, whereas others are shorter rating scales that have parent-, teacher-, and self-report forms. The range of ages for these instruments extends from 18 months up to adulthood, with some of them split into different versions based on age groupings. The time to complete the assessments can range from 5 minutes for a simple parental questionnaire up to 150 minutes for a more complicated assessment given by a trained professional, such as the Autism Diagnostic Interview-Revised (ADI-R).

The ADI-R is a standardized, semi-structured investigator-based interview for caregivers of children and adults for whom autism is a possible diagnosis, which provides a diagnostic algorithm for the International Classification of Disease (ICD), tenth edition definition of autism and the Diagnostic and Statistical Manual of Mental Disorders, fourth edition (DSM-IV). It includes 93 items focusing on Early Development, Language and Communication, Reciprocal Social Interactions, and Restricted, Repetitive Behaviors and Interests \cite{ADIR:2003}. 

The Autism Diagnostic Observation Schedule, Second Edition (ADOS-2) is a standardized protocol for the observation of social and communicative behaviors of persons with autism and related disorders. The instrument consists of a series of structured and semistructured prompts for interaction, accompanied by the coding of specific target behaviors associated with particular tasks and by general ratings of the quality of behaviors. The ADOS-2 consists of four modules so that each one is appropriate for children and adults at different developmental and language levels \cite{ADOS:2012}.

The Behavior Assessment for Children, Third Edition (BASC-3) is a norm-referenced diagnostic tool that uses a multi-dimensional approach to assess the behavior and self-perceptions of children and young adults ages 2 through 25 years. The BASC-3 includes teacher and parent rating scales separated into three forms: preschool, child and adolescent, as well as a self-report of personality separated into four forms: interview, child, adolescent, and college.  It includes 23 clinical, adaptive and content scales, as well as ten clinical and executive functioning indexes and five composite scores that can be used to assist with differential diagnoses when used in conjunction with the Diagnostic and Statistical Manual of Mental Disorders, fifth edition (DSM-V) \cite{BASC:2015}.

The Behavior Rating Inventory of Executive Function, Second Edition (BRIEF2) is an informant-report rating scale designed to assess executive behaviors in children and adolescents. It consists of nine domains of executive functioning, combined into three summary indexes, including the Behavioral Regulation Index (BRI), the Emotional Regulation Index (ERI), and the Cognitive Regulation Index (CRI). The BRI captures the child's ability to regulate and monitor behavior effectively, while the ERI represents the child's ability to regulate emotional responses and to shift set or adjust to changes in the environment, people, or plans. The CRI then reflects the child's ability to control and manage cognitive processes and to problem solve effectively. The BRIEF2 can be used in conjunction with other rating scales, clinical interviews and observations to diagnose children and adolescents who have either developmental or acquired neurological conditions, such as learning disabilities, attention disorders, traumatic brain injuries, and other medical conditions \cite{BRIEF:2015}.

The Child Behavior Checklist (CBCL) is a parent-report questionnaire for evaluating maladaptive behavioral and emotional problems in children and adolescents aged 2 to 18. It assesses a wide-range of internalizing behaviors, such as anxiety and depression, as well as externalizing behaviors, such as aggression and hyperactivity. When used in conjunction with the other rating scales within the Achenbach System of Empirically Based Assessment (ASEBA), the teacher-report and self-report questionnaires, it can be used to assess six DSM-V diagnostic categories, including Depressive Problems, Attention Deficit/Hyperactivity Problems, Anxiety Problems, Oppositional Defiant Problems, Somatic Problems, and Conduct Problems \cite{CBCL:2001}.

The Conners, 3rd Edition (Conners 3) is a thorough assessment of ADHD and its most commonly associated problems and disorders in school-aged youth. It is a multi-informant assessment with forms for parents, teachers, and youth. The assessment features multiple content scales that assess ADHD-related concerns as well as related problems in executive functioning, learning, aggression, and peer/family relations. In addition to these content scales, Conners 3 has five DSM-IV Symptom Scales that can be used as diagnostic criteria for ADHD and common comorbid disorders, including ADHD Inattentive, ADHD Hyperactive-Impulsive, ADHD Combined, Conduct Disorder, and Oppositional Defiant Disorder scales \cite{Conners:2008}.

The Social Responsiveness Scale-Second Edition (SRS-2) is a 65-item rating scale measuring deficits in social behavior associated with ASD, as outlined by the DSM-IV. The SRS-2 consists of four rating forms across three age ranges, including parent-, teacher-, and self-report forms. There are five treatment sub-scales, including Social Awareness, Social Cognition, Social Communication, Social Motivation, and Restricted Interests and Repetitive Behavior, as well as an overall total score that are used to assess ASD \cite{SRS:2012}.

The American Academy of Pediatrics (AAP) and the National Initiative for Children's Healthcare Quality (NICHQ) jointly published the Vanderbilt ADHD Diagnostic Rating Scale (VADRS) as a psychological assessment toolkit to be used in the assessment and treatment of ADHD in a primary care setting. It includes versions specific for parents and teachers. In addition to items corresponding to the ADHD diagnostic criteria of the DSM-IV, the VADRS includes symptom screens for four common comorbidities: oppositional defiant disorder, conduct disorder, anxiety, and depression \cite{Vanderbilt:2003}.

\begin{table}[htbp!]
\caption{Child Behavioral Instruments}
\centering
\begin{tabular}{p{1.75cm} p{2.5cm} p{2.25cm} p{3cm} p{1.25cm} p{3.75cm}}
\hline\hline
\multicolumn{1}{l}{\textbf{Instrument}} & \multicolumn{1}{l}{\textbf{Age Range}} & \multicolumn{1}{l}{\textbf{Completion Time}} & \multicolumn{1}{l}{\textbf{Reporter(s)}} & \multicolumn{1}{l}{\textbf{Number of Items}} & \multicolumn{1}{l}{\textbf{Conditions Covered}} \\
\hline
ADI-R \cite{ADIR:2003} & 2 years - Adult & 90-150 minutes & Clinician & 93 & Autism \\
ADOS-2 \cite{ADOS:2012} & 12 months - Adult & 40-60 minutes & Clinician & 28-38 & Autism \\
BASC-3 \cite{BASC:2015} & 2-25 years & 10-30 minutes & Parent, Teacher, Self & 105-192 & Autism, ADHD, Anxiety, Conduct Disorder, and \\
 &  &  &  &  & Depressive Disorders\\
BRIEF 2 \cite{BRIEF:2015} & 5-18 years & 5-30 minutes & Parent, Teacher, Self & 55-63 & Autism, ADHD, Learning \\
 &  &  &  &  & disabilities, and other \\
 &  &  &  &  & acquired neurological \\
  &  &  &  &  & conditions \\
CBCL \cite{CBCL:2001} & 1.5-18 years & 10-30 minutes & Parent, Teacher, Self & 100-113 & ADHD, Anxiety, Conduct \\
 &  &  &  &  & Problems, Depression, \\
 &  &  &  &  & Oppositional Defiant, and \\
 &  &  &  &  & Somatic Problems \\
Conners 3 \cite{Conners:2008} & 6-18 years & 5-20 minutes & Parent, Teacher, Self & 99-115 & ADHD, Conduct Disorder, and Oppositional Defiant Disorder\\
SRS-2 \cite{SRS:2012} & 2.5 years - Adult & 15-20 minutes & Parent, Teacher, Self & 65 & Autism \\
VADRS \cite{Vanderbilt:2003} & 6-12 years & 5-20 minutes & Parent, Teacher & 43-55 & ADHD, Anxiety, Conduct Disorder, Depression, and Oppositional Defiant \\
 &  &  &  &  & Disorder \\
\hline \hline
\end{tabular}
\label{tab:Instruments}
\end{table}

\section*{Methods}

We set out to build the first generation of project Rosetta to include a minimal set of Rosetta questions representative of all of the concepts identified within the ontology. Eight child behavioral instruments were included in the analysis that led to the creation of the behavioral ontology underlying Rosetta, which were described in more detail above. The process for creating Rosetta involved ingesting the existing child behavioral instruments into a consistent format, creating Rosetta questions and answer choices, and mapping existing instrument questions to Rosetta questions. The process for building Rosetta is detailed in this section.

\subsection*{Procedure}
\subsubsection*{Ingestion of Instruments}

The first step of this process involved creating a document to ingest each of the eight child behavioral instruments into a consistent format, including the ADI-R, ADOS-2, BASC-3, BRIEF2, CBCL, Conners 3, SRS-2, and the VADRS. Each instrument was included in a single tab within the document for reference named by the version of the instrument. All versions of each instrument by age group were included, whereas only the parental version of an instrument was included if there were multiple forms for different raters. A uniform set of column names was used for all instruments, including question id, question body, and answer choices. This was carried out to have all of the instruments together in a standard format to make it easier to compare the concepts being asked in each instrument to create a comprehensive child behavioral ontology. 

\subsubsection*{Clinical Domain Categorization}

The individual questions from the instruments that were ingested in the previous step were added to an aggregate tab within the document. This made it easier to examine each question to determine which questions from these instruments had overlapping semantic meaning. Since there were a total of 1274 questions from all of the versions of the existing instruments, the questions needed to be grouped into categories to be able to assess this semantic overlap. A scheme was created to use for categorizing each of the individual questions into an ontology of clinical domains, shown in Table \ref{tab:Domains}. A team of subject matter experts which included clinical neuropsychology, developmental psychology, and pediatric neurology representations were consulted in order to arrive at groupings and sub-groupings within the ontology. At the top level, there were three broad domains, including Cognitive, Motor, and Somatic, which were then further broken down into a total of 60 leaf categories.

Following the creation of the ontology, each question was assigned to a leaf category within the ontology. Through this process of categorization, the goal was to be able to have a workable amount of existing questions within each of the leaf categories to better understand the specific concepts being asked in each domain.
\newpage

\begin{center}
\begin{longtable}[htbp!]{p{3cm} p{3cm} p{4cm} p{4cm}}
\caption{Clinical Domain Categories} \\
\hline \hline \multicolumn{1}{l}{\textbf{Level 1}} & \multicolumn{1}{l}{\textbf{Level 2}} & \multicolumn{1}{l}{\textbf{Level 3}} & \multicolumn{1}{l}{\textbf{Level 4}} \\
\hline
\endfirsthead

\multicolumn{4}{c}%
{{\bfseries \tablename\ \thetable{} -- continued from previous page}} \\
\hline \hline \multicolumn{1}{l}{\textbf{Level 1}} & \multicolumn{1}{l}{\textbf{Level 2}} & \multicolumn{1}{l}{\textbf{Level 3}} & \multicolumn{1}{l}{\textbf{Level 4}} \\ \hline 
\endhead

\hline \multicolumn{4}{r}{{Continued on next page}} \\ 
\endfoot

\hline \hline
\endlastfoot

\hline
Cognitive & Behavioral & Emotional & Adaptability \\
 &	&  & Anger Control \\
 &	&  & Anxiety \\
 &	&  & Depression \\
 &	&  & Mood \\
 &	&  & Obsessive Compulsive \\
 &	&  & Paranoia \\
Cognitive & Behavioral & Sensory & Disturbed \\
 &	&  & Intrigued \\
Cognitive & Behavioral & Social & Aggression \\
 &	&  & Atypicality \\
 &	&  & Awareness \\
 &	&  & Comforting \\
 &	&  & Conduct \\
 &	&  & Ego \\
 &	&  & Eye Contact \\
 &	&  & Facial Expressions \\
 &	&  & Group Play \\
 &	&  & Imitation \\
 &	&  & Joint Attention \\
 &	&  & Leadership \\
 &	&  & Maturity \\
 &	&  & Reciprocal Interaction \\
 &	&  & Relationships \\
 &	&  & Shared Interests \\
 &	&  & Smile \\
 &	&  & Staring \\
 &	&  & Withdrawal \\
Cognitive & Executive & Attention &  \\
 & Functioning & Confusion & \\
 &  & Coping & \\
 &  & Fluency & \\
 &  & General & \\
 &  & Hyperactivity & \\
 &  & Imagination & \\
 &  & Impulsivity & \\
 &  & Inhibitory Control & \\
 &  & Memory & \\
 &  & Patience & \\
 &  & Perseveration & \\
 &  & Planning & \\
 &  & Reasoning & \\
Cognitive & Language and & Expressive &  \\
 & Communication & Nonverbal Communication & \\
 &  & Receptive & \\
 &  & Speech & \\
Motor & Fine &  &  \\
 & Gross & Lower &  \\
 &  & Upper & \\
Somatic & Dermatologic &  &  \\
 & Fatigue &  & \\
 & Gastrointestinal &  & \\
 & General &  & \\
 & Illness &  & \\
 & Neurologic &  & \\
 & Sleep &  & \\
 & Vision &  & \\
 & Weight &  & \\
\label{tab:Domains}
\end{longtable}
\end{center}

\subsubsection*{Rosetta Question Creation}

Following the categorization of questions, the questions were grouped by leaf category to determine which questions from the existing instruments were conceptually similar and therefore, could be covered by the creation of a novel single Rosetta question. As the questions were analyzed by leaf category, Rosetta questions were phrased to ensure a minimal loss of meaning to assess each particular child behavior.  Subject matter experts drafted de novo questions based on the features identified (i.e., the broad clinical domain and leaf categories). Question versions were then reviewed to arrive at a final wording that was both novel as well as conceptually similar to the original questions.  As an example of this iterative process, we looked at the Adaptability leaf category and found there were 32 existing questions within this leaf from six different instruments. A subset of these questions are shown in Table \ref{tab:Questions}, to illustrate how well these questions overlap between instruments. 

\begin{table}[htbp!]
\caption{Sample Questions Within Adaptability Leaf Category}
\centering
\begin{tabular}{p{3.5cm} p{8.5cm}}
\hline\hline
\multicolumn{1}{l}{\textbf{Instrument}} & \multicolumn{1}{l}{\textbf{Question}} \\
\hline
ADI-R & 74. Is bothered by minor changes in routine, schedule or how personal things are arranged \\
ADI-R & 75. Gets upset by changes around the house \\
BASC-3 (Preschool) & 88. Adjusts well to new surroundings \\
BASC-3 (Child) & 47. Adjusts well to changes in plans \\
BASC-3 (Adolescent) & 156. Adjusts well to change in teacher \\
BRIEF 2 & 11. Has trouble adjusting to new situations \\
CBCL & 21. Disturbed by changes in routine \\
SRS-2 & 24. Difficulty with changes to routine \\
\hline \hline
\end{tabular}
\label{tab:Questions}
\end{table}

A single Rosetta question was created by a team of subject matter experts to assess a child’s ability to adapt to changes to a routine, schedule or the environment.  Again, we utilized a process of experts phrasing a de novo question followed by a team review and final phrasing. This particular Rosetta question was phrased as follows: “Does [NAME] become unusually upset with or have difficulty accepting small changes? For example, a change in [his/her] bedtime routine, weekly scheduled activities, or furniture arrangement in the house.” 

\subsubsection*{Rosetta Question Mapping}

A mapping was then created in a separate tab in the document so that each of the original instrument questions were mapped to a corresponding Rosetta question. As shown in the example above, all of the sample questions in Table \ref{tab:Questions} could  be mapped to the Rosetta question about a child's ability to adapt to changes. Within the adaptability leaf category, twenty-one instrument questions were mapped to this Rosetta question. Three additional Rosetta questions were created within this leaf category to assess other specific child behaviors associated with adaptability, and the remaining eleven existing questions within the adaptability leaf category were mapped accordingly. This mapping process led to a many-to-one mapping, where many questions from different instruments  mapped to a single Rosetta question. On average, three instruments and seven questions mapped to one Rosetta question. Further details of the overlap created by carrying out this process are discussed in the Results section.

\subsubsection*{Rosetta Question Answer Creation and Mapping}

As multiple questions mapped to a single Rosetta question, each of these existing questions tended to have varying types of answer responses. The ADI-R and ADOS-2 generally have descriptive answer choices that relate to the quality of behavior being assessed, whereas the remaining assessments have answer choices on a Likert scale referring to the frequency of that behavior. For the last step in this process, these differences had to be consolidated to create a new, consistent coding of answer choices that each of the original question responses could be mapped to, retaining as many of the response signals as possible.

\begin{table}[htbp!]
\caption{Answer Choices for Questions Mapped to Routine Change Question}
\centering
\begin{tabular}{p{1.5cm} p{3cm} p{3cm} p{3.5cm} p{4.5cm}}
\hline\hline
\multicolumn{1}{l}{\textbf{Instrument}} & \multicolumn{1}{l}{\textbf{Answer 1}} & \multicolumn{1}{l}{\textbf{Answer 2}} & \multicolumn{1}{l}{\textbf{Answer 3}} & \multicolumn{1}{l}{\textbf{Answer 4}} \\
\hline
ADI-R & No difficulties with & Unusually negative & Definite, unusual & Definite, unusual resistance to \\
& changes to routine & reaction to minor & reactions to minor & minor changes, with impairment \\
 &  & changes & changes, causing distress & of family activities \\
 & & & & \\
BASC-3 & Never & Sometimes & Often & Almost Always \\
BRIEF2 & Never & Sometimes & Often & \\
CBCL & Not True & Somewhat True & Very True \\
SRS-2 & Not True & Sometimes True & Often True & Almost Always True \\
\hline \hline
\end{tabular}
\label{tab:Answers}
\end{table}

Each of the original questions and answer choices was independently examined to make sure the answer codes were mapped without a significant loss of meaning. In the example discussed in the section above, five different instruments were mapped to the question about adaptability to change and each of them was asked in a slightly different way with different answer choices. The corresponding ADI-R questions had four descriptive answer choices, whereas the BRIEF2 and CBCL had three answer choices on a frequency scale, and the BASC-3 and SRS-2 had four answer choices on a frequency scale, shown in Table \ref{tab:Answers}. 

\begin{figure}[H]
\centering
\includegraphics[width=\linewidth]{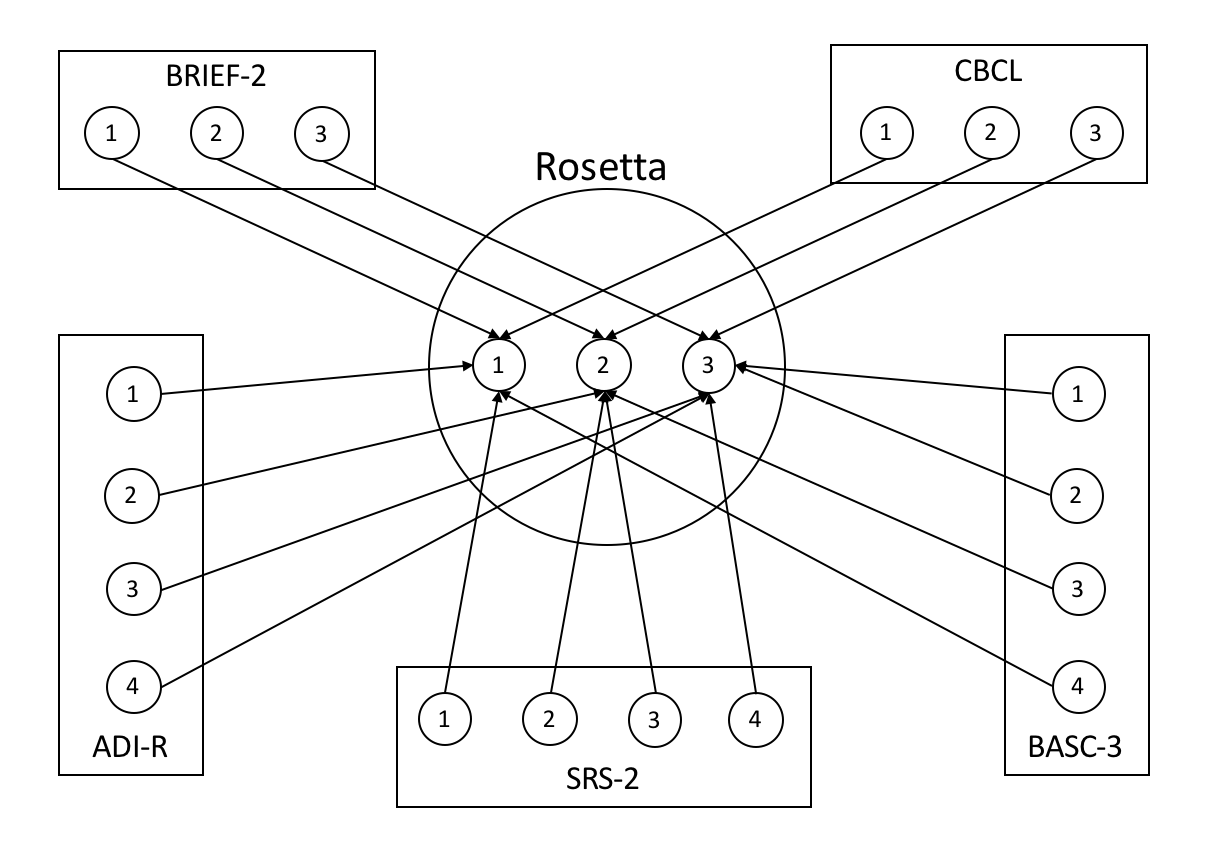}
\caption{Instrument answer choices mapped to Rosetta answer codes for Adaptability, Routine Change question}
\label{fig:Mapping}
\end{figure}

The subject matter experts crafted question answer choices for Rosetta questions such that, where appropriate, descriptive quality-based responses of ADI-R and ADOS were combined with the frequency responses typical of instruments like BASC-3 and BRIEF2. When questions from BRIEF2 or CBCL mapped to a Rosetta question, three answer codes were created in Rosetta because that was the least amount of responses that would potentially be mapped if the child only had the BRIEF2 or CBCL instruments assessed. This was decided because it could not be inferred how a parent would have responded if given more answer choices. The new answer choices for this particular question that combined frequency and quality were as follows: 1=Rarely or never; 2=Sometimes, but with little interference in family life; and 3=Often, and with some interference with family life. We then mapped each of the existing question answer choices to the Rosetta answer codes based on how the questions and answer choices overlapped with the phrasing of the Rosetta question and answer codes as shown in Figure \ref{fig:Mapping}.

\subsubsection*{Case study: machine-learning-based, concurrent assessment of children for autism and ADHD}

To demonstrate the utility of Rosetta as a platform to enable the development of simultaneous multi-condition assessment algorithms, we trained and validated a machine-learning algorithm to assess young children for the risk of autism or ADHD using a single questionnaire comprised entirely of Rosetta questions. The input data for this case study consisted of 3,731 patient records of children aged 4-10, each of whom underwent one or more clinical assessment instruments that are currently incorporated into Rosetta. The diagnostic labels for the dataset were assigned by licensed medical professionals, and the breakdown was 2,941 positive for autism, 343 positive for ADHD, and 447 negative for both.

As is the case in most clinical data collection settings, no single assessment instrument has been undergone by all patients in the dataset for this case study. Rather, the data covers 6 different Rosetta-friendly clinical assessment instruments with little overlap. Because many instruments have multiple questionnaire versions, the total number of unique instrument questionnaire versions was 15. Under traditional settings, it would not be possible to proceed with machine learning training in these conditions. However, with Rosetta available, it was possible to leverage the entire dataset as training and cross-validation samples to a machine-learning predictive algorithm.

The assessment algorithm identifies autism and ADHD using the Rosetta dataset as follows: first, a data imputation technique is used to infer values for missing Rosetta questions for every sample as needed. Next, a gradient boosted decision tree algorithm is trained to identify if either of autism or ADHD is present for a child. A second gradient boosted decision tree algorithm is trained to identify which of the two conditions is present, and is only used for predictions if the first algorithm identifies a child as having autism or ADHD. To train each algorithm an iterative procedure is used to identify the most predictive Rosetta questions to be used in model training. The training process identified a total of 30 Rosetta questions as the relevant features for the assessment of autism and ADHD in a single questionnaire.

Cross-validation AUC is evaluated to be 99\% (when identifying autism or ADHD) and 99\% (for separating autism from ADHD). This encouraging preliminary result demonstrates the potential utility of the application, and the benefit of applying Rosetta instrument mappings to unlock the power of machine-learning in settings that might otherwise not be amenable to such application. Further clinical trial testing should be performed to evaluate how effective such algorithms are when the Rosetta instrument is applied in real world settings.

\newpage

\section*{Results}

The final Rosetta document mapped 1274 existing instrument questions to 209 Rosetta questions. Table \ref{tab:Resulting_fusion} shows the resulting fusion of questions, including the number of existing instruments with overlap, the total existing instrument questions and resulting number of novel Rosetta questions for each leaf category in the ontology. On average, three instruments and seven existing questions mapped to a single Rosetta question.

\begin{center}
\begin{longtable}[htbp!]{p{2.5cm} p{3.5cm} p{1.75cm} p{1.75cm} p{1.75cm}}
\caption{Resulting Fusion of Instrument Questions within each Leaf Category} \\
\hline \hline \multicolumn{1}{l}{\textbf{Base Category}} & \multicolumn{1}{l}{\textbf{Leaf Category}} & \multicolumn{1}{l}{\textbf{Overlapping Instruments}} & \multicolumn{1}{l}{\textbf{Instrument Questions}} & \multicolumn{1}{l}{\textbf{Rosetta Questions}} \\
\hline
\endfirsthead

\multicolumn{5}{c}%
{{\bfseries \tablename\ \thetable{} -- continued from previous page}} \\
\hline \hline \multicolumn{1}{l}{\textbf{Base Category}} & \multicolumn{1}{l}{\textbf{Leaf Category}} & \multicolumn{1}{l}{\textbf{Overlapping Instruments}} & \multicolumn{1}{l}{\textbf{Instrument Questions}} & \multicolumn{1}{l}{\textbf{Rosetta Questions}} \\ \hline 
\endhead

\hline \multicolumn{5}{r}{{Continued on next page}} \\ 
\endfoot

\hline \hline
\endlastfoot

\hline
Cognitive, & Adaptability & 6 & 32 & 4 \\
Behavioral, & Anger Control & 6 & 48 & 5 \\
Emotional & Anxiety & 7 & 126 & 17 \\
 & Depression & 4 & 38 & 5 \\
 & Mood & 4 & 16 & 3 \\
 & Obsessive Compulsive & 4 & 15 & 4 \\
 & Paranoia & 3 & 6 & 1 \\
 & Emotional & 5 & 35 & 8 \\
\hline
Cognitive, & Disturbed & 2 & 4 & 1 \\
Behavioral, & Intrigued & 3 & 6 & 1 \\
Sensory & Sensory & 3 & 16 & 3 \\
\hline
Cognitive, & Aggression & 6 & 57 & 3 \\
Behavioral, & Atypicality & 3 & 32 & 6 \\
Social & Awareness & 8 & 52 & 11 \\
& Comforting & 3 & 5 & 1 \\
& Conduct & 4 & 100 & 14 \\
& Ego & 1 & 2 & 1 \\
& Eye Contact & 5 & 11 & 1 \\
& Group Play & 4 & 10 & 2 \\
& Imitation & 2 & 3 & 1 \\
& Joint Attention & 3 & 22 & 3 \\
& Leadership & 1 & 7 & 1 \\
& Maturity & 1 & 3 & 1 \\
& Reciprocal Interactions & 2 & 22 & 2 \\
& Relationships & 4 & 32 & 4 \\
& Shared Interests & 4 & 16 & 4 \\
& Smile & 2 & 2 & 1 \\
& Staring & 3 & 7 & 1 \\
& Withdrawal & 5 & 51 & 4 \\
& Social & 7 & 32 & 8 \\
\hline
Cognitive, & Attention & 6 & 62 & 7 \\
Executive & Confusion & 2 & 2 & 1 \\
Functioning & Coping & 1 & 9 & 1 \\
& Fluency & 2 & 6 & 3 \\
& General & 1 & 9 & 7 \\
& Hyperactivity & 7 & 34 & 4 \\
& Imagination & 3 & 11 & 1 \\
& Impulsivity & 5 & 20 & 3 \\
& Inhibitory Control & 3 & 7 & 2 \\
& Memory & 5 & 14 & 3 \\
& Patience & 4 & 7 & 1 \\
& Perseveration & 5 & 18 & 1 \\
& Planning & 4 & 24 & 4 \\
& Reasoning & 3 & 18 & 4 \\
& Executive Functioning & 5 & 18 & 6 \\
\hline
Cognitive, & Expressive & 4 & 77 & 12 \\
Language \& & Nonverbal & 2 & 10 & 2 \\
Communication & Receptive & 6 & 17 & 5 \\
& Speech & 4 & 12 & 3 \\
\hline
Motor & Fine & 2 & 8 & 1 \\
 & Gross & 4 & 11 & 4 \\
\hline
Somatic & Dermatologic & 1 & 2 & 1 \\
 & Fatigue & 3 & 8 & 1 \\
 & Gastrointestinal & 2 & 18 & 2 \\
 & General & 2 & 15 & 1 \\
 & Illness & 1 & 10 & 1 \\
 & Neurologic & 2 & 10 & 2 \\
 & Sleep & 1 & 7 & 2 \\
 & Vision & 1 & 2 & 1 \\
 & Weight & 1 & 1 & 1 \\
 & Somatic & 1 & 2 & 1 \\
\label{tab:Resulting_fusion}
\end{longtable}
\end{center}

\begin{figure}[htbp!]
\centering
\includegraphics[width=\linewidth]{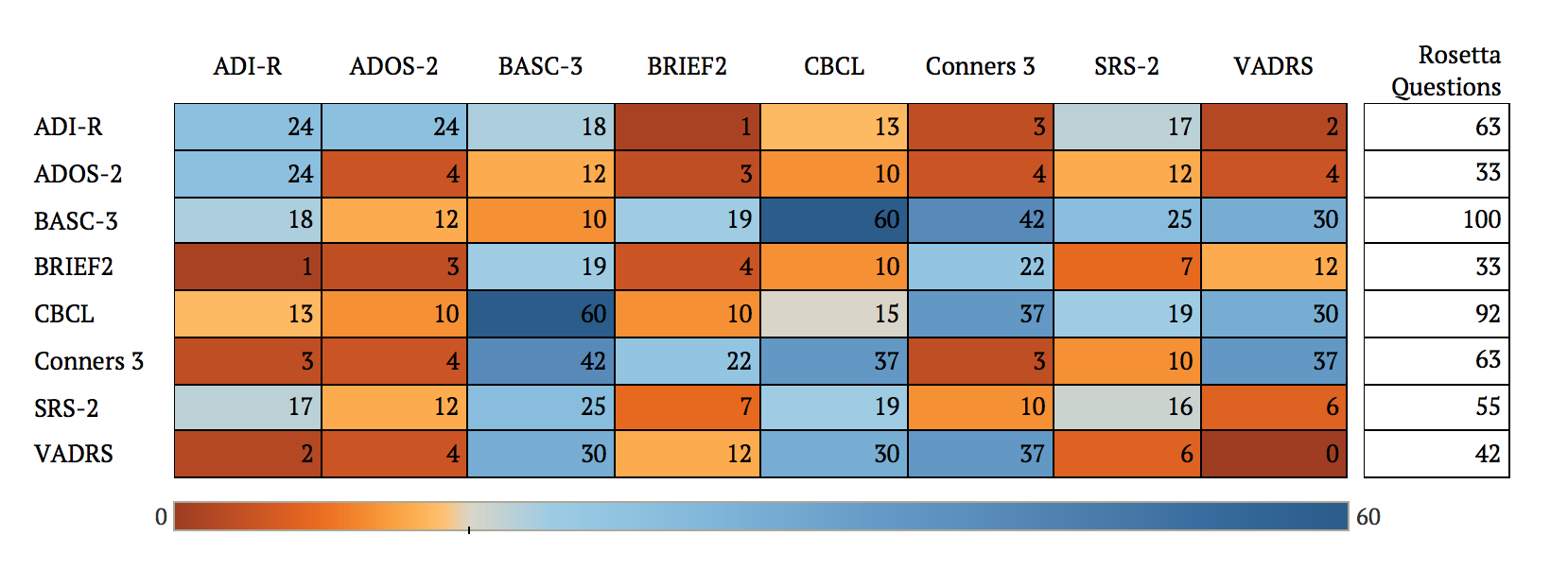}
\caption{Heat map showing the overlap between instruments for all Rosetta questions.}
\label{fig:Heatmap}
\end{figure}

\newpage
The resulting overlap between existing instruments and Rosetta questions is illustrated in the heat map in Figure \ref{fig:Heatmap}. For each existing instrument, the heat map shows how many Rosetta questions have overlap with every other instrument in the ontology. The diagonal from top left to bottom right shows how many Rosetta questions only map to one instrument. It can be seen from this that the ADI-R, CBCL, and SRS-2 have a considerable amount of questions that are difficult to find overlap with other existing questions. The last column in this figure shows the resulting number of Rosetta questions that overlap with each instrument.

\section*{Discussion}

To our knowledge, our first generation of project Rosetta is the first ontological mapping of its kind, creating a minimal set of questions with significant conceptual overlap across multiple childhood behavioral/developmental instruments. The resulting ontology incorporates many concepts that have diagnostic relevance for child behavioral conditions, which can be used as a resource for child mental, behavioral, and developmental health diagnosis and treatment. By creating a reduced set of 209 questions, Rosetta can be used to create more concise instruments, thereby addressing  some of the time constraints that lead to delays in diagnosis and treatment interventions. 

The overlap between existing childhood behavioral/developmental instruments that was created by Rosetta can be used to create a virtual diagnostic instrument that covers more patients across various ages and with various conditions in a uniform way that could not be done before. This ability to take in and combine assessment data from any existing instrument through the corresponding mappings allows for the creation of a large, dense dataset that is required for building machine learning algorithms in the development of diagnostic tools as presented in our case study in the Methods section above.  

There are some potential limitations to this project due to the variation between the instruments included in the first generation of Rosetta. There could be a loss of response signals from over-simplification of the question phrasing when creating the Rosetta questions, as well as from mapping existing instrument questions to Rosetta questions that are not representative of a particular behavior. Another challenge leading to a loss of response signals comes from the combination of instruments with varying scales of answer choices, as well as the combination of descriptive quality-based answer choices with frequency-based answer choices. Both of these limitations could lead to a misrepresentation of parental responses for particular behaviors. 

This work needs to be extended to cover more child behavioral health instruments. Different child behavioral health instruments could potentially expand the ontology to be more representative of other diagnoses that are not well-represented by the eight instruments included in this ontology. This work could also be extended into other diagnostic domains, such as adult behavioral conditions by applying the same concepts to adult checklists and screening tools. Additionally, clinical trial testing should be performed to assess the application of the Rosetta instrument in real world settings across a variety of child behavioral conditions.

\newpage
\bibliography{references}

\begin{thebibliography}{10}
\urlstyle{rm}
\expandafter\ifx\csname url\endcsname\relax
  \def\url#1{\texttt{#1}}\fi
\expandafter\ifx\csname urlprefix\endcsname\relax\def\urlprefix{URL }\fi
\expandafter\ifx\csname doiprefix\endcsname\relax\def\doiprefix{DOI: }\fi
\providecommand{\bibinfo}[2]{#2}
\providecommand{\eprint}[2][]{\url{#2}}

\bibitem{CDC:2016}
\bibinfo{author}{Bitsko, R.~H.} \emph{et~al.}
\newblock \bibinfo{journal}{\bibinfo{title}{Health care, family, and community
  factors associated with mental, behavioral, and developmental disorders in
  early childhood - united states, 2011-2012}}.
\newblock {\emph{\JournalTitle{MMWR Morbidity and Mortality Weekly Report}}}
  \textbf{\bibinfo{volume}{65}}, \bibinfo{pages}{221--26}
  (\bibinfo{year}{2016}).

\bibitem{Screening:2017}
\bibinfo{author}{Kuhn, C.} \emph{et~al.}
\newblock \bibinfo{journal}{\bibinfo{title}{Effective mental health screening
  in adolescents: Should we collect data from youth, parents or both?}}
\newblock {\emph{\JournalTitle{Child Psychiatry \& Human Development}}}
  \textbf{\bibinfo{volume}{48}}, \bibinfo{pages}{385--92}
  (\bibinfo{year}{2017}).

\bibitem{ADIR:2003}
\bibinfo{author}{Rutter, M.}, \bibinfo{author}{LeCouteur, A.} \&
  \bibinfo{author}{Lord, C.}
\newblock \emph{\bibinfo{title}{Autism Diagnostic Interview - Revised}}
  (\bibinfo{publisher}{Western Psychological Services}, \bibinfo{address}{Los
  Angeles, CA, USA}, \bibinfo{year}{2003}).

\bibitem{ADOS:2012}
\bibinfo{author}{Lord, C.}, \bibinfo{author}{Rutter, M.},
  \bibinfo{author}{DeLavore, P.~C.} \& \bibinfo{author}{Risi, S.}
\newblock \emph{\bibinfo{title}{Autism Diagnostic Observation Schedule, Second
  Edition}} (\bibinfo{publisher}{Western Psychological Services},
  \bibinfo{address}{Los Angeles, CA, USA}, \bibinfo{year}{2012}).

\bibitem{BASC:2015}
\bibinfo{author}{Reynolds, C.~R.} \& \bibinfo{author}{Kamphaus, R.~W.}
\newblock \emph{\bibinfo{title}{Behavior Assessment System for Children, Third
  Edition}} (\bibinfo{publisher}{Pearson Education, Inc.},
  \bibinfo{address}{Bloomington, MN, USA}, \bibinfo{year}{2015}).

\bibitem{BRIEF:2015}
\bibinfo{author}{Gioia, G.~A.}, \bibinfo{author}{Isquith, P.~K.},
  \bibinfo{author}{Guy, S.~C.} \& \bibinfo{author}{Kenworthy, L.}
\newblock \emph{\bibinfo{title}{Behavior Rating Inventory of Executive
  Function, Second Edition}} (\bibinfo{publisher}{PAR}, \bibinfo{address}{Lutz,
  FL, USA}, \bibinfo{year}{2015}).

\bibitem{CBCL:2001}
\bibinfo{author}{Achenbach, T.~M.} \& \bibinfo{author}{Rescorla, L.~A.}
\newblock \emph{\bibinfo{title}{Manual for the ASEBA School-Age Forms \&
  Profiles}} (\bibinfo{publisher}{University of Vermont, Research Center for
  Children, Youth, \& Families}, \bibinfo{address}{Burlington, VT, USA},
  \bibinfo{year}{2001}).

\bibitem{Conners:2008}
\bibinfo{author}{Conners, C.~K.}
\newblock \emph{\bibinfo{title}{Conners Comprehensive Behavior Rating Scales
  Manual}} (\bibinfo{publisher}{Multi-Health Systems},
  \bibinfo{address}{Toronto, Ontario, Canada}, \bibinfo{year}{2008}).

\bibitem{SRS:2012}
\bibinfo{author}{Constantino, J.~N.}
\newblock \emph{\bibinfo{title}{The Social Responsiveness Scale, Second
  Edition}} (\bibinfo{publisher}{Western Psychological Services},
  \bibinfo{address}{Los Angeles, CA, USA}, \bibinfo{year}{2012}).

\bibitem{Vanderbilt:2003}
\bibinfo{author}{Wolraich, M.~L.} \emph{et~al.}
\newblock \bibinfo{journal}{\bibinfo{title}{Psychometric properties of the
  vanderbilt adhd diagnostic parent rating scale in a referred population}}.
\newblock {\emph{\JournalTitle{Journal of Pediatric Psychology}}}
  \textbf{\bibinfo{volume}{8}}, \bibinfo{pages}{559--67}
  (\bibinfo{year}{2003}).

\end{thebibliography}




\end{document}